\documentclass[runningheads]{llncs}
 \makeatletter
\def\@fnsymbol#1{\ensuremath{\ifcase#1\or \or *\or \ddagger\or
   \mathsection\or \mathparagraph\or \|\or **\or \dagger\dagger
   \or \ddagger\ddagger \else\@ctrerr\fi}}
    \makeatother
\usepackage[T1]{fontenc}
\usepackage{graphicx}
\usepackage{tabularx}
\usepackage{booktabs}
\usepackage{amsmath}
\usepackage{xcolor}
\usepackage{subcaption}

\usepackage{textcomp}

\begin{document}
\title{Semi-supervised Learning using Robust Loss} %Divergence\thanks{Supported by organization x.}}
\author{Wenhui Cui$^*$, Haleh Akrami$^*$\thanks{$^*$Authors have equal contribution}, Anand A. Joshi, Richard M. Leahy}
%
%\authorrunning{***** et al.}
\institute{University of Southern California}
\maketitle              % typeset the header of the contribution
\begin{abstract}
The amount of manually labeled data is limited in medical applications, so semi-supervised learning and automatic labeling strategies can be an asset for training deep neural networks. However, the quality of the automatically generated labels can be uneven and inferior to manual labels. In this paper, we suggest a semi-supervised training strategy for leveraging both manually labeled data and extra unlabeled data. In contrast to the existing approaches, we apply robust loss for the automated labeled data to automatically compensate for the uneven data quality using a teacher-student framework. First, we generate pseudo-labels for unlabeled data using a teacher model pre-trained on labeled data. These pseudo-labels are noisy, and using them along with labeled data for training a deep neural network can severely degrade learned feature representations and the generalization of the network. Here we mitigate the effect of these pseudo-labels by using robust loss functions. Specifically, we use three robust loss functions, namely beta cross-entropy, symmetric cross-entropy, and generalized cross-entropy. We show that our proposed strategy improves the model performance by compensating for the uneven quality of labels in image classification as well as segmentation applications.

\end{abstract}
\section{Introduction}
Deep neural networks usually require a large amount of labeled training data to achieve good performance. However, manual annotations, especially for medical images, are very time-consuming and costly to acquire. So it is desirable to incorporate extra knowledge from unlabeled data into the training process and assist supervised training. The dominant methods that leverage unlabeled data for classification, specifically semantic segmentation, include (1) consistency training \cite{french2019semi,ke2019dual} that ensure consistency of prediction in the presence of various perturbations. In these approaches, a standard supervised loss term (e.g., cross-entropy loss) is combined with an unsupervised consistency loss term that enforces consistent predictions in response to perturbations applied to unsupervised samples; and (2) pseudo-labeling \cite{zhu2020improving,zou2018unsupervised,li2019bidirectional,zou2019confidence,9003388} of the unlabeled images obtained from the model trained on the labeled images by generating pseudo-labels using another or even the same neural network. 
%for generating pseudo labels, self-training is one possible strategy.
Generating pseudo-labels is a straightforward and effective way to enrich supervised information~\cite{9003388}.
However the pseudo-labeled datasets inevitably include mislabeled data that introducing noise. These weakly labeled data can have a disproportionate impact on the learning process, and the model may over-fit to the outliers \cite{zhang2021understanding}.
 
A major challenge in using auto-labeled data for training is accounting for noise in the pseudo-labels \cite{ren2020not,zhu2020improving,zou2018unsupervised,li2019bidirectional,zou2019confidence}. Several approaches deal with noise in pseudo-labeled data by focusing on heuristically controlling their use by (1) lowering the ratio of pseudo-labels in each mini-batch \cite{zou2018unsupervised}; (2) selecting pseudo-labels with high confidence \cite{zou2019confidence}; or (3) setting lower weights in computing the loss for pseudo-labels \cite{ren2020not}.

Alternative approaches that mitigate the effect of noisy labels can be categorized into three classes: (1) label correction methods that improve the quality of raw labels by modeling characteristics of the noise and correcting incorrect labels \cite{xiao2015learning}; (2) methods with robust loss that are inherently robust to labeling errors \cite{wang2019symmetric}; and (3) refined adaptive training strategies that are more robust to noisy labels \cite{yu2019does}. 

Here we focus on robust loss functions that offer a theoretically-based approach to the noisy label problem \cite{wang2019symmetric}. Previous studies have shown some loss functions such as Mean Absolute Error (MAE) that were originally designed for regression problems can also be used in classification settings \cite{ghosh2017robust}. However, training with MAE has been found to be very challenging because of the gradient saturation issue \cite{zhang2018generalized} . The Generalized Cross-Entropy (GCE) \cite{zhang2018generalized} loss applies a Box-Cox transformation to probabilities that has been shown to be a generalized mixture of MAE and Cross-Entropy (CE). Using a similar idea of applying a power-law function, beta-cross entropy loss has been developed to mitigate the effect of noise in the training data \cite{akrami2020robust,akrami2022robust}. Minimizing beta-cross entropy (BCE) is equivalent to minimizing beta-divergence\cite{basu1998robust}, which is the robust counterpart of KL-divergence. BCE has an extra normalization term compared to the GCE loss. Another study suggested Symmetric Cross-Entropy (SCE) loss by combining Reverse Cross-Entropy (RCE) together with the CE loss \cite{wang2019symmetric}.

Here we develop a semi-supervised learning strategy to utilize labeled data and weakly labeled data. A teacher-student training framework~\cite{xie2020self,9003388} is adopted. We propose to first generate pseudo-labels of the unlabeled data using a teacher model trained on ground-truth labels. Then we train a student model using a combination of ground-truth and pseudo-labels. We apply a robust loss to enhance model robustness to noise in pseudo-labels, so that both supervised and weakly supervised knowledge are combined in an optimized learning strategy. We demonstrate the effectiveness of the proposed strategy on a simple classification task and a brain tumor segmentation task.

In this work, our contribution is three-fold:
\begin{itemize}
    \item We improved model performance when only limited labeled data is available, which is especially meaningful to medical images. % or
    \item We proposed a simple yet effective semi-supervised learning strategy by introducing a plug-and-play module: the robust loss function.
    \item The proposed strategy is agnostic to specific model architecture and can be applied to various segmentation or classification or even regression tasks.
\end{itemize}

\section{Materials and Methods}
First, we introduce the robust loss functions used for handling noise in pseudo-labels in Sec.\ref{subsec:robust_loss_func} and illustrate the utility of the robust loss functions using a simulation. We then describe our semi-supervised strategy in Sec.\ref{sec:semi}, which employs a teacher-student framework \cite{xie2020self}. 
%We demonstrate the effectiveness of our method on a standard classification dataset (CIFAR-10) and a medical image segmentation dataset.

In a multi-class classification setting, the most commonly used loss function is multivariate cross-entropy given by:
\begin{align}
\label{eq:ce}
    \mathcal{L}_{CE}=& \sum_{k=1}^{k} -q(k|x)log(p(k|x)) 
\end{align}
where $x$ is the input variable, $y$ is the response variable. $p(k|x)$ is the probability output of a \textit{deep neural network}~(DNN) classifier, and $q(k|x)$ is the one-hot encoding of the label. We use this as a baseline in our experiments below, however this loss is function is susceptible to label noise, which we address with the robust approaches described below.  

\begin{figure}
\centering
\includegraphics[width=0.8\textwidth]{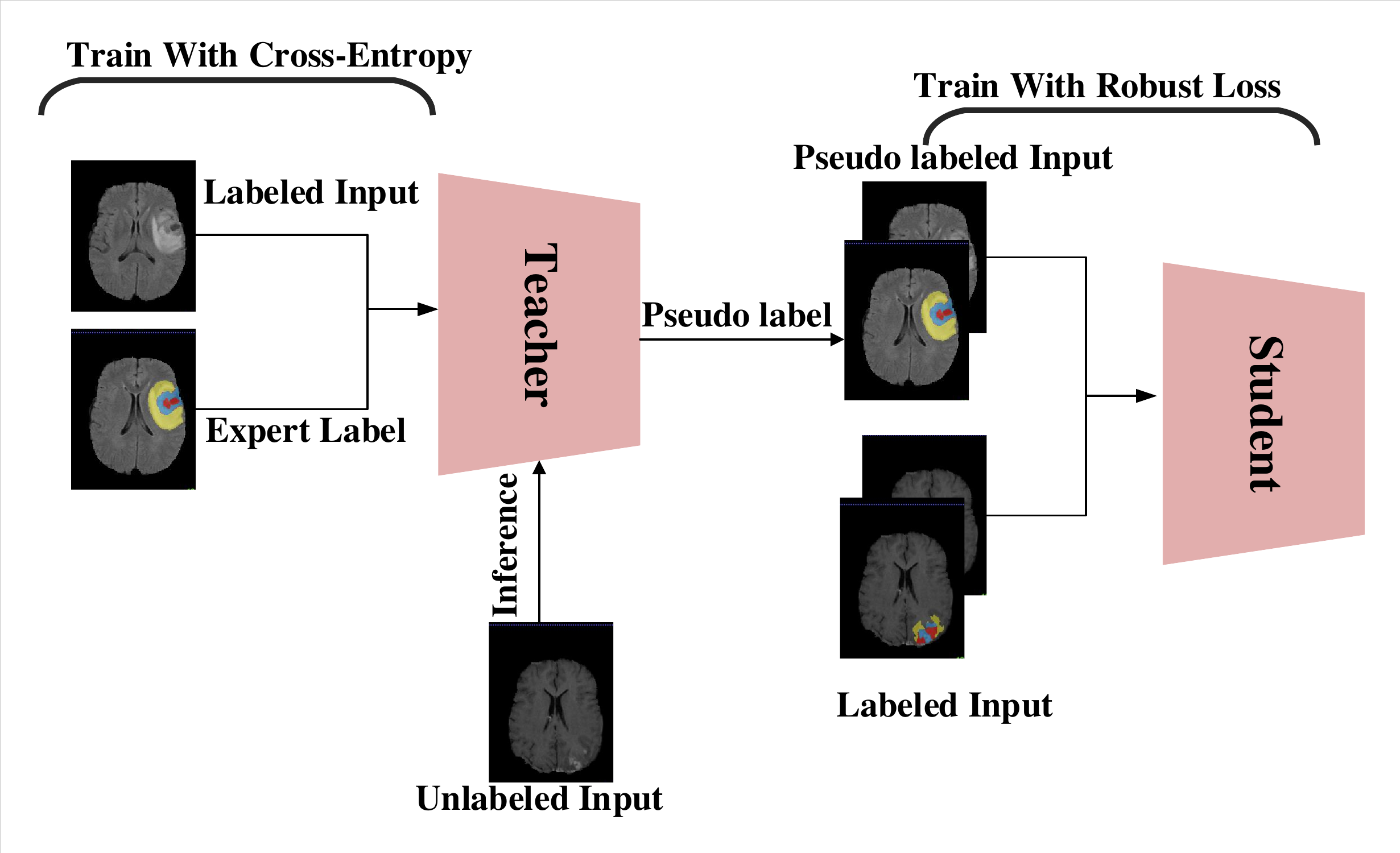}
\caption{Framework: We develop a teacher-student semi-supervised framework by using both manually labeled as well as pseudo-labeled data. We propose to first generate pseudo-labels of the unlabeled data using a model trained on manually labeled ground-truth labels. Then we train a second model using these ground-truth labels and generated pseudo-labels simultaneously by applying a robust loss to enhance model robustness to noise in the pseudo-labels.} \label{fig:framework}
\end{figure}

\subsection{Robust loss functions}
\label{subsec:robust_loss_func}
We evaluated three different robust loss functions to reduce the effect of pseudo-label noise.
The Generalized Cross-Entropy (GCE) loss is defined as follows:
\begin{align}
    \mathcal{L}_{GCE}=\frac{(1-p(y|x))^q}{q}
\end{align}
 GCE applies a Box-Cox transformation to probabilities (power-law function). Using L’Hôpital’s rule it can be shown that GCE is equivalent to CE for $\lim q \to 0$ and to MAE when $\lim q \to 1$ so this loss is a generalization of CE and MAE \cite{zhang2018generalized}.

The Beta Cross-Entropy (BCE) loss can be expressed as: 
\begin{equation}
\label{eq:bce}
    \mathcal{L}_{BCE}=\frac{{\beta+1}}{\beta} (1-p(y|x))^\beta+ \sum_{k=1}^{k} p(k|x)^{\beta+1}
\end{equation}

BCE minimizes $\beta$-divergence \cite{basu1998robust} between the posterior and empirical distributions when the posterior is a categorical distribution \cite{akrami2020brain,akrami2022robust}. Using L’Hôpital’s rule, it can be shown that BCE is equivalent
to CE for $\lim \beta \to 0$ where $\beta$-divergence also converges to KL-divergence. $\beta$-divergence is the robust counterpart of KL-divergence  using a power-law function. BCE has an extra regularization term compared to GCE. This loss has not previously been applied to classification tasks \cite{akrami2020robust,akrami2022robust}.    

The Reverse Cross-Entropy (RCE) loss is defined as \cite{wang2019symmetric}:
\begin{align}
    \mathcal{L}_{RCE}=& \sum_{k=1}^{k} -p(k|x)log(q(k|x)) \\
    =& -p(y|x)log1-\sum_{k\neq y}p(k|x)A \nonumber\\
    =& -A\sum_{k\neq y}p(k|x)A=-A(1-p(y|x) \nonumber
\end{align}

$A$ is the smoothed/clipped replacement of $\log \{0\}$. MAE is a special case of RCE at $A = -2$. RCE has also been proved to be robust to label
noise, and can be combined with CE to obtain the Symmetric
Cross-Entropy (SCE): $\mathcal{L}_{SCE}=\alpha CE+\gamma RCE$~\cite{wang2019symmetric}, where $\alpha$ and $\gamma$ are hyper-parameters. 

\begin{figure}[!t]
\centering
\includegraphics[width=0.7\textwidth]{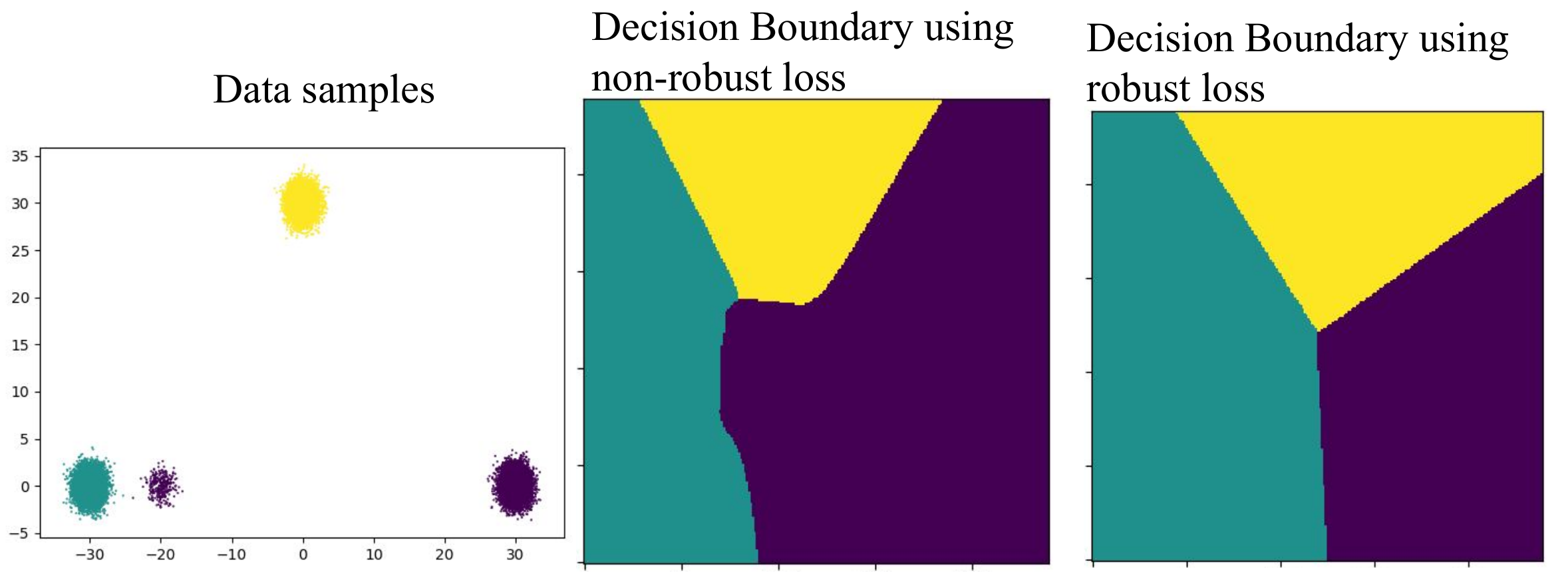}
\caption{An illustrative example of using robust loss for classification: (left) a simulated dataset with three distinct classes and a mislabeled subclass data, shown as a smaller cluster on the bottom left; (b) decision boundary computed using a single layer perceptron and multivariate cross-entropy loss (non-robust loss); (c) the decision boundary calculated using a single layer perceptron, but with multivariate beta-cross entropy (robust loss). It can be seen that the decision boundary computed using the non-robust loss is affected due to the mislabeled outliers, whereas the decision boundaries calculated using a robust loss are minimally impacted by the mislabeled data.} \label{fig:robust_illustration}
\end{figure}

As an illustrative example (Figure \ref{fig:robust_illustration}), we created a simulated dataset with three classes using Gaussian mixtures plus outliers. The figure shows color-coded data according to their classes. A single layer perceptron was trained with the multivariate cross-entropy loss function (Eq. \ref{eq:ce}), which is a non-robust loss. The decision boundary determined by the network can be seen to be impacted by the presence of outliers in the data. The procedure was repeated by training the perceptron with multivariate $\beta$-cross entropy loss (Eq. \ref{eq:bce}), which is a robust loss. It can be seen that the decision boundaries, in this case, are minimally impacted by the presence of outliers. 

\subsection{Training using unlabeled data and a robust loss function} \label{sec:semi}
We introduce a teacher-student framework~\cite{xie2020self} to perform semi-supervised learning for classification and its special case semantic segmentation. We assume we have a small number of labeled training samples and a large set of extra unlabeled samples. We first train a teacher model with standard cross-entropy loss using the labeled set and then generate pseudo-labels for the unlabeled set. The quality of pseudo-labels depends on the performance and generalizability of the teacher model. When the number of labeled samples is smaller, the generated pseudo-labels will become noisier. We then combine the true and pseudo-labels to  re-train a student model.
We used CE for human-annotated labels and the robust loss functions for pseudo-labels. 
An illustration of our framework is shown in Fig.~\ref{fig:framework}.

\section{Experiments and Results}

To explore the effectiveness of our semi-supervised strategy, we performed image classification and segmentation tasks. We started with a simple classification task using the CIFAR-10 \cite{krizhevsky2009learning} dataset. We used this dataset as robust loss functions were previously applied on this dataset \cite{ma2020normalized} using simulated noisy labels, and we wanted to investigate if these loss functions are useful for the semi-supervised learning using the same dataset. Then we performed a brain tumor segmentation task on the BraTS 2018 dataset~\cite{menze2014multimodal,Bakas2017,bakas2018identifying} to further explore the benefits of our proposed strategy for medical imaging applications.
To create a semi-supervised experiment setting, we first divide the training set into two groups, assuming a specific percentage ($p$) of subjects being the labeled group and the rest of subjects being the unlabeled group. Denote the total number of training subjects as $N$.
%, Denote the number of subjects as $N$. 
We define performance of the teacher model trained on $N*p$ subjects as the lower bound and the model trained on $N$ subjects (the entire training set) as the upper bound. We generated the pseudo-labels of $N*(1-p)$ subjects using the teacher model trained on the other $N*p$ subjects. Then a student model is trained on both ground-truth labels and pseudo-labels.
% In our experiments, we add different subjects to the labeled and unlabeled groups, and generate different combinations of two groups by changing the value of $p$. 

\subsection{CIFAR-10}
   First, we performed experiments on the CIFAR-10 dataset~\cite{krizhevsky2009learning}  (with 50000 images for training and 10000 for testing). We set $p=10\%$ and trained our teacher network with ground truth labels. 
   Then we generated pseudo-labels for the rest of 90\% of the data and combined this set with the 10\% ground truth labels to train the student model with CE, GCE, SCE, and BCE. The hyper-parameters for GCE, SCE and BCE losses were chosen based on a validation set (10\% of the training data ($\beta=5$), ($q=0.9$), ($\alpha=0.1$, $\gamma=0.01$)). For both teacher and student networks we used ResNet-18~\cite{he2015deep} and trained the networks for 150 epochs using  stochastic gradient descent \cite{robbins1951stochastic} (momentum=0.9, weight decay=1e-4). The initial learning rate was 0.1 and reduced to 0.01 after 100 epochs.
    The results are summarized in Table \ref{tab:cfar}. All robust losses improved the model performance.

%\subsection{Skin Lesion Segmentation}   
%In this experiment we trained a U-net model~\cite{ronneberger2015unet} to perform skin lesion segmentation using ISIC 2017 dataset~\cite{codella2018skin}. This dataset consists of 2000 images for training, and 600 for testing \cite{codella2018skin}. We used half of the test set as the validation set to tune the hyper-parameters of the robust loss functions ($\beta=5),$($q=0.9)$,($\alpha=0.1, \beta=0.1$). We resized the images to 128*128.
%In this experiment, we set $p=5\%$, meaning that we assume the rest $95\%$ of the data is unlabeled and train these data using pseudo labels. 

%Our result demonstrates the effectiveness of our training strategy and using a robust loss(Table \ref{tab1}). The Dice coefficient using the entire data-set with ground truth labels (upper bound) was 0.795 and reduced to 0.625 by only using 0.05 of the data. We could increase this value to 0.6833 with our training strategy. The learning rate was 0.001; we used RMSProp \cite{hinton2012neural} optimizer and trained the network for 50 epochs. We ran the experiments three times to get a mean and standard deviation for the Dice coefficient. 

\begin{table}[!t]
\caption{Performance of proposed semi-supervised strategy on CIFAR-10 dataset. Results show test accuracy of the models trained using only 10\% of ground truth labels. }\label{tab1}
%The second row shows the Dice coefficient(mean and standard deviation) for semi-supervised training using ISIC 2017 data-set using only 5\% of ground truth labels.
%\resizebox{0.98\textwidth}
\centering
%\resizebox{0.98\textwidth}{h}{
\begin{tabular}{|c|c|c|c|c|c|c|}
\hline
Data-set & Lower bound&CE&BCE&GCE  &SCE &Upper bound\\
\hline
CIFAR10 (Accuracy) & $66.17\%$ & $69.62\%$ & $79.53\%$ & $77.68\%$ & $78.74\%$ & $89.31\%$ \\ 
\hline
\end{tabular}
%}
\label{tab:cfar}
\end{table}

%ISIC 2017 (Dice coeff.) & 0.6250() & 0.6350() & 0.6833() & 0.6713() & 0.6614() & 0.7950() 
\subsection{Brain Tumour Segmentation}
\textbf{Backbone Model: }
We adopted a 3-dimensional CNN called TransBTS~\cite{wang2021transbts} as our backbone model, which combines U-Net~\cite{ronneberger2015unet} and Transformer~\cite{vaswani2017attention} networks. 
TransBTS is based on an encoder-decoder structure and takes advantage of Transformer to learn not only local context information but also global semantic correlations~\cite{wang2021transbts}. 
TransBTS achieved superior performance compared to previous state-of-the-art models for the brain tumor segmentation task~\cite{wang2021transbts}. 
We trained the model for $1000$ epochs from scratch with an Adam optimizer. The initial learning rate is $0.0002$ and the batch size is set to $4$.\\
%implementation details?
\textbf{Dataset and Evaluation: }
We performed experiments on a publicly available dataset provided by the Brain Tumor Segmentation (BraTS) 2018 challenge~\cite{menze2014multimodal,Bakas2017,bakas2018identifying}. 
The BraTS 2018 Training dataset includes $285$ subjects with ground-truth labels. 
%And each subject has four image modalities: T1, T1Gd, T2, and FLAIR, all of which are used at the same time during model training.
The Magnetic Resonance Images~(MRIs) have been registered into a common space and the image dimension is $240\times240\times 155$.
The ground-truth labels contain three tumor tissue classes (necrotic and non-enhancing tumor: label 1, peritumoral edema: label 2, and GD-enhancing tumor: label 4) and background: label 0.
We split the $285$ subjects into a training set (200 subjects), a validation set (28 subjects), and a test set (57 subjects).
We used the validation set to select the values of hyper-parameters, which are $\beta=0.001$ for BCE loss; $q=0.7$ for GCE loss; $\alpha=0.01$, $\gamma=1.0$ for SCE loss.
We used Dice score to quantitatively evaluate the segmentation accuracy for enhancing tumor region (ET, label 1), regions of the tumor core (TC, labels 1 and 4), and the whole tumor region (WT, labels 1, 2, and 4). 
To explore the benefits of our strategy when different numbers of ground-truth labels are available, we evaluated the segmentation accuracy when $p=30\%, 50\%, 70\%$ of data is considered as labeled, respectively.  We ran the experiments three times to compute mean and standard deviation for the Dice score, each time we select different subjects as labeled data.

\begin{figure}[htb!]
\centering
\begin{subfigure}[b]{0.32\textwidth}
    \centering
    \includegraphics[width=\linewidth]{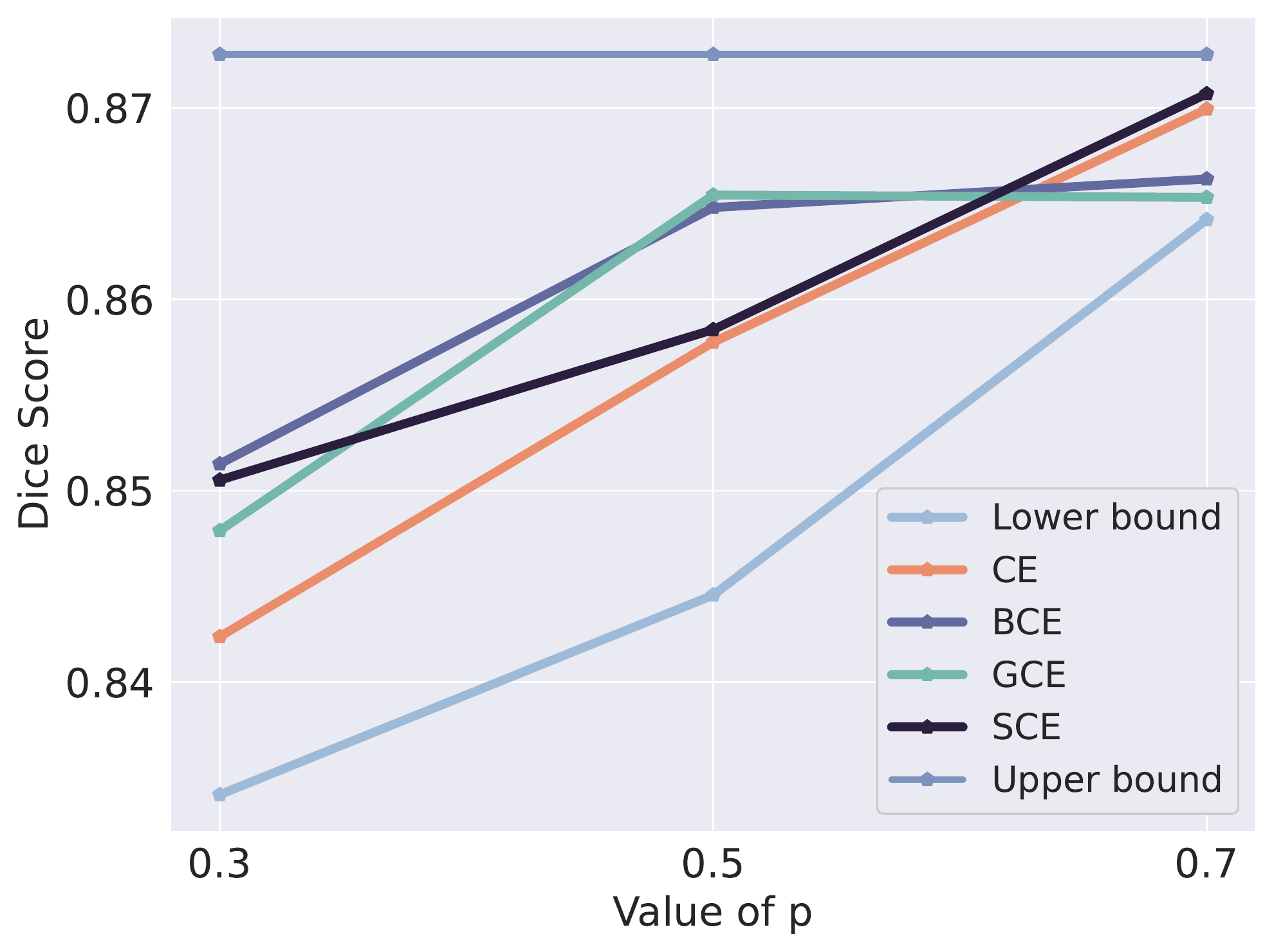}
    \caption{WT} \label{fig:dice_coeffs1}
\end{subfigure}
\begin{subfigure}[b]{0.32\textwidth}
    \centering
    \includegraphics[width=\linewidth]{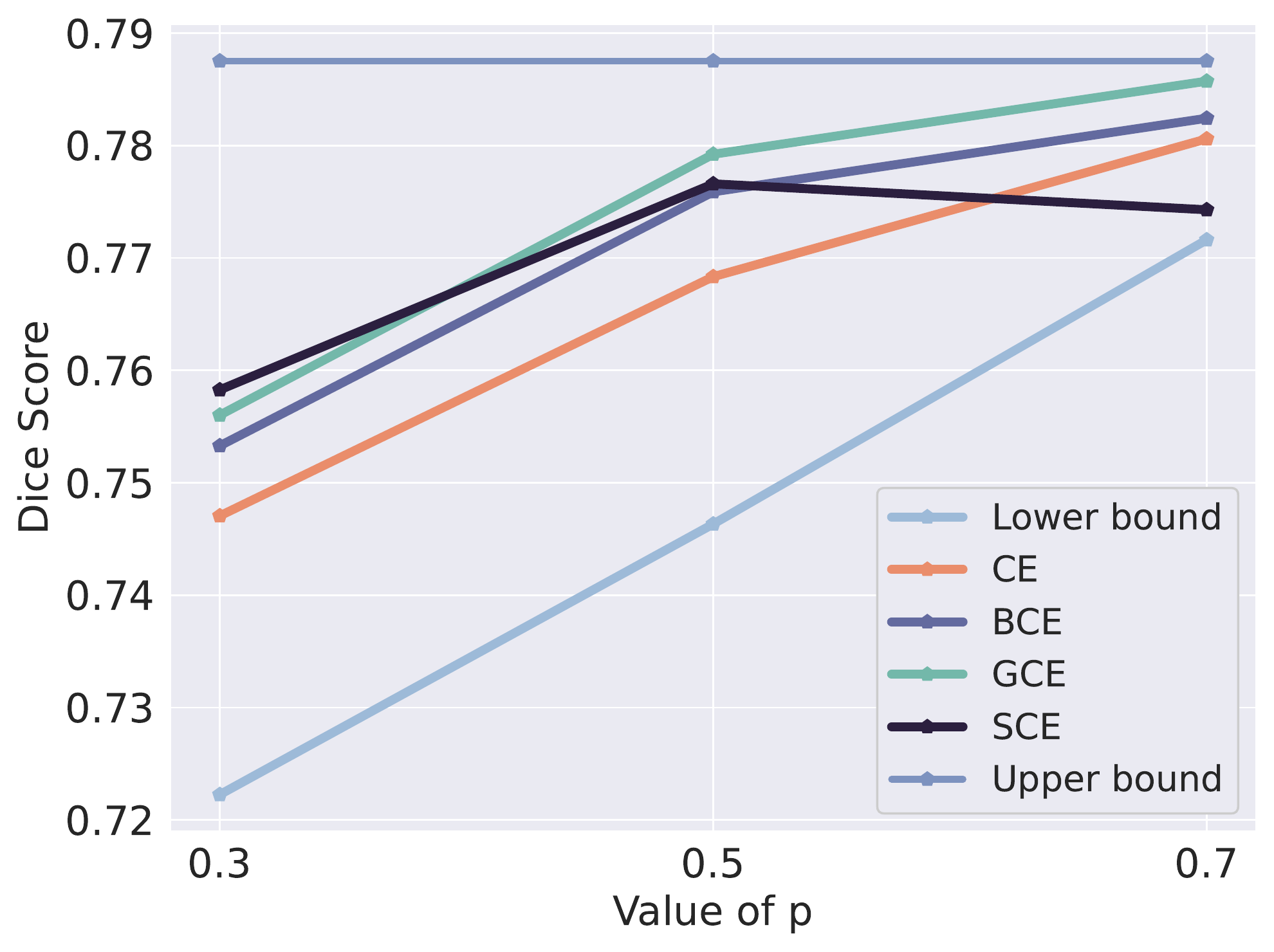}
    \caption{TC} \label{fig:dice_coeffs2}
\end{subfigure}
\begin{subfigure}[b]{0.32\textwidth}
    \centering
    \includegraphics[width=\linewidth]{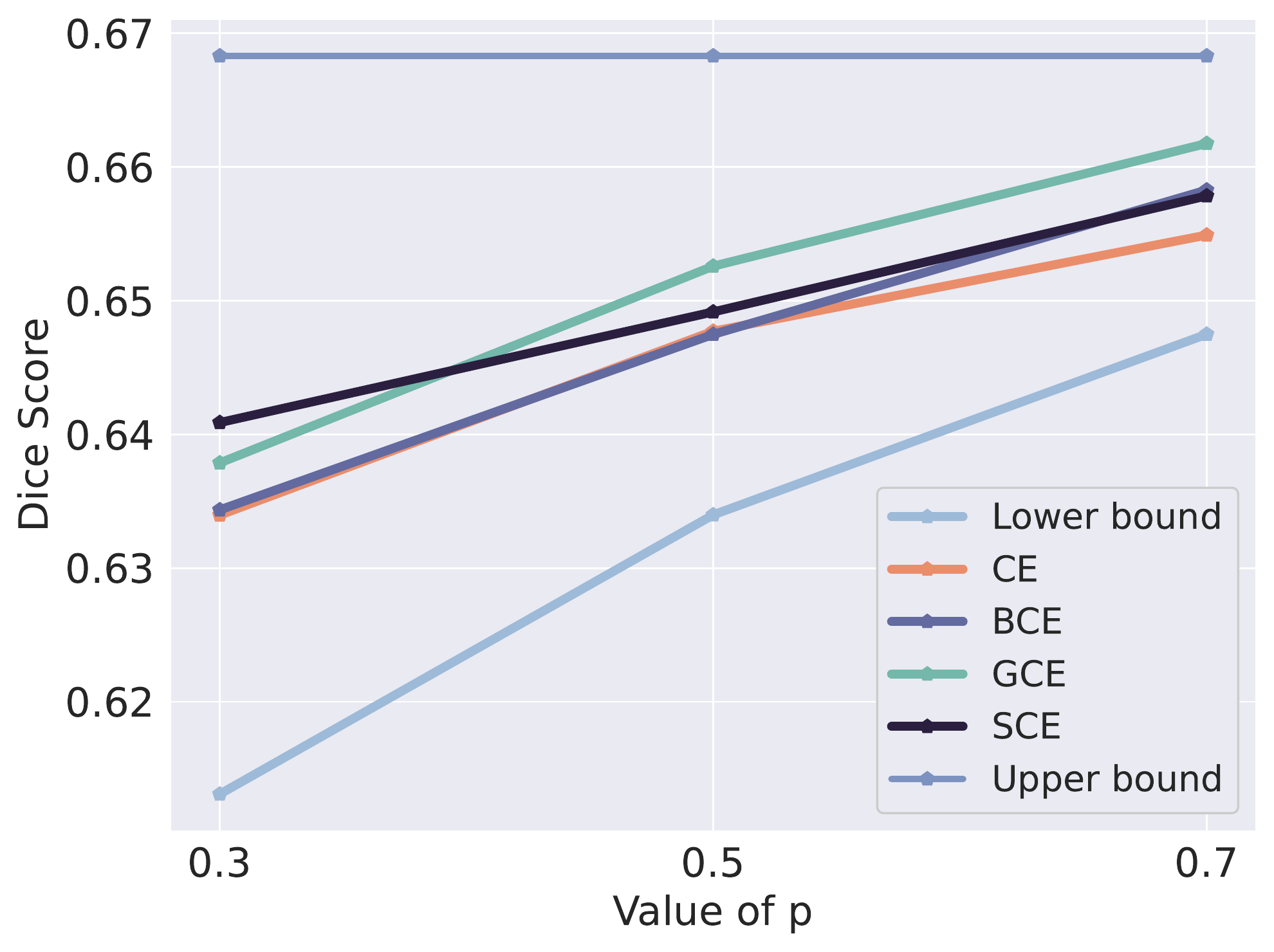}
    \caption{ET} \label{fig:dice_coeffs3}
\end{subfigure}

\caption{A graphical illustration of segmentation accuracy improvement using the proposed strategy for different fractions of pseudo-labels ($p$) in the training data. Figures (a), (b) and (c) show the average dice scores of tumor classes WT, TC, and ET, respectively.}
\label{fig:dices}
\end{figure}

\begin{figure}[htbp!]
\centering
\includegraphics[width=0.96\textwidth]{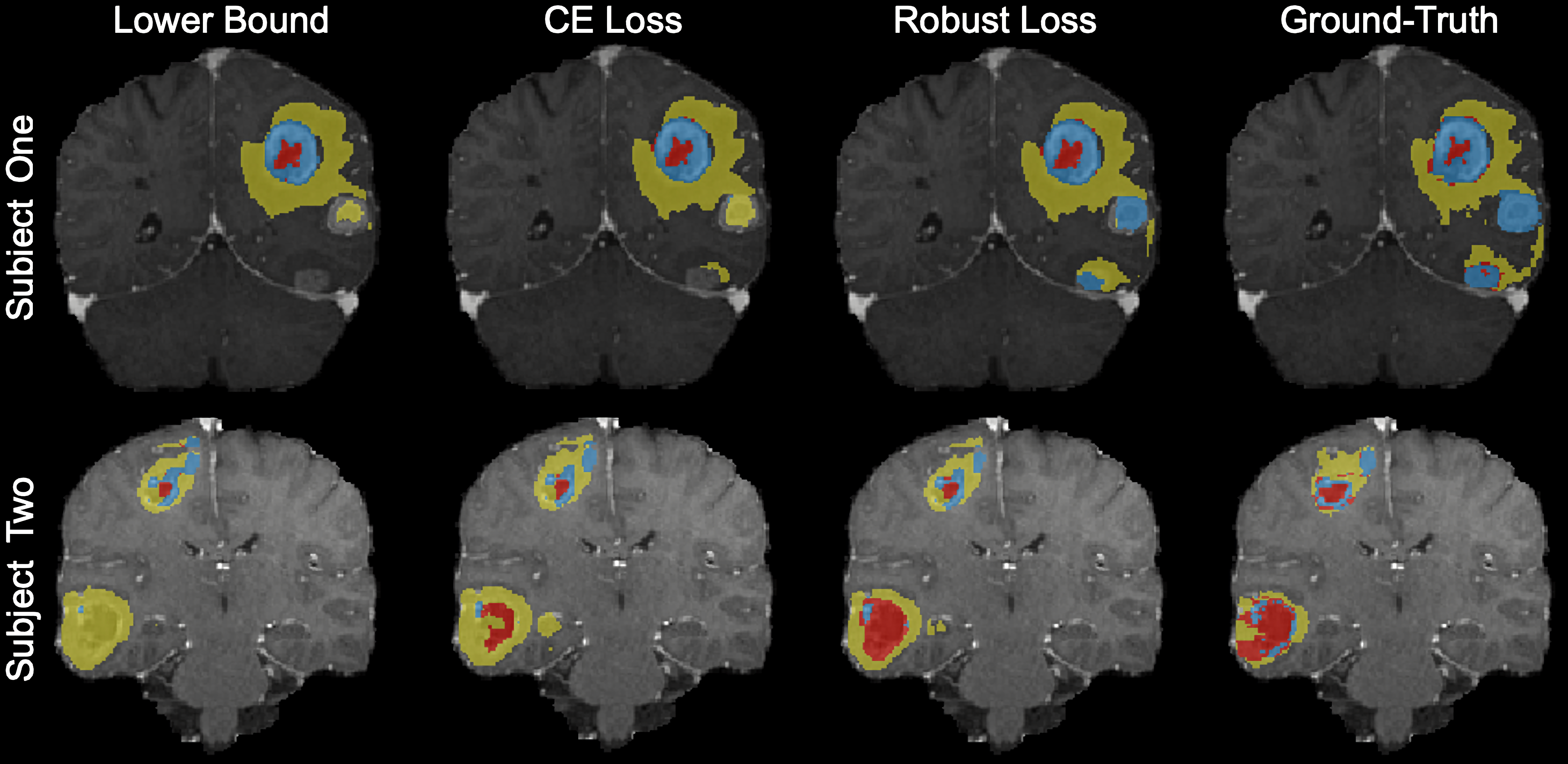}
\caption{Comparison of Brain Tumor Segmentation Results: from left to right showing the segmentation results of the lower bound model, the model with CE loss, the model with robust loss, and the ground-truth label. Segmentation results indicate label 1 (yellow), label 2 (blue) and label 4 (red), where we have ET (label 1), TC (labels 1 and 4), and WT ( labels 1, 2, and 4). } 
\label{fig:seg}
\end{figure}

Table~\ref{tab:dsc} compares the dice scores of segmentation results when different loss functions (CE, BCE, GCE, SCE) are applied during training. From Table~\ref{tab:dsc} and Fig.~\ref{fig:dices} we can observe that applying robust loss improves the segmentation accuracy compared to the CE loss and the lower bound. The improvement is more significant when only a small number of labeled data is available. This is because a relatively large proportion of noisy labels has a more negative effect on the model performance, and applying robust loss can make the model less perturbed by noise. The proposed strategy at $p=30\%, p=50\%$ achieved even better performance than the lower bound model at $p=50\%, p=70\%$, and comparable performance compared to the model with CE loss at $p=50\%, p=70\%$.
When $p=70\%$, some of the robust loss results showed slightly worse dice scores compared to CE loss for the WT and TC classes. This is probably because the teacher model generated higher quality pseudo-labels when more ground truth labels are available, the noise level is negligible and these two classes are relatively easier to segment, so adding robust loss will not further boost model performance. To qualitatively evaluate the segmentation results, we selected two representative test subjects and showed the segmentation results produced by different approaches as well as ground-truth labels in Fig.~\ref{fig:seg}. 
Evidently, segmentation results generated by the model with robust loss are more accurate, which verified the benefits of our semi-supervised strategy.
\begin{table}[t]
\caption{Comparison of the mean and standard deviation of Dice scores for different methods on test subjects for different tumor classes (WT, TC, ET).}
\centering
\resizebox{0.99\textwidth}{!}{

% \begin{tabular}{|c|*{18}{c|}}  % repeats {c|} 18 times
% \hline
% \multicolumn{1}{|c|}{ }  & \multicolumn{9}{c|}{Dice Score} & \multicolumn{9}{c|}{Hausdorff Distance ($95\%$)} \\ \cline{2-19}
% \multicolumn{1}{|c|}{Methods}  & \multicolumn{3}{c|}{WT} & \multicolumn{3}{c|}{TC} & \multicolumn{3}{c|}{ET} & 
% \multicolumn{3}{c|}{WT} & \multicolumn{3}{c|}{TC} & \multicolumn{3}{c|}{ET} \\ \cline{2-19} 
% \multicolumn{1}{|c|}{} & $30\%$ & $50\%$ & $70\%$ &$30\%$ & $50\%$ & $70\%$ &$30\%$ & $50\%$ & $70\%$ &$30\%$ & $50\%$ & $70\%$ &$30\%$ & $50\%$ & $70\%$ & $30\%$ & $50\%$ & $70\%$ \\ \hline
%  Lowerbound & 0.8341 \pm & 0.8446 \pm 0.0052 & & 0.7223 \pm & 0.7463\pm0.0109 & &0.6131 \pm & 0.6340\pm 0.0044 & & 9.0169 \pm & 8.1761\pm0.2722 & & 8.3402 \pm & 7.5551\pm0.1677 & & 3.6843 &  3.6942\pm0.0257 &  \\ 
%  CE & & & & & & & & & & & & & & & & &  \\ 
%  BCE & & & & & & & & & & & & & & & & &  \\ 
%  GCE & & & & & & & & & & & & & & & & &  \\ 
%  SCE & & & & & & & & & & & & & & & & &  \\ 
%  Upperbound & & & & & & & & & & & & & & & & &  \\ 
% \end{tabular}
\begin{tabular}{|c|*{9}{c|}}  % repeats {c|} 18 times
\hline
\multicolumn{1}{|c|}{ }  & \multicolumn{9}{c|}{Dice Score}  \\ \cline{2-10}
\multicolumn{1}{|c|}{Methods}  & \multicolumn{3}{c|}{WT} & \multicolumn{3}{c|}{TC} & \multicolumn{3}{c|}{ET} \\ \cline{2-10} 
\multicolumn{1}{|c|}{} & 30\% & 50\% & 70\% & 30\% & 50\% & 70\% & 30\% & 50\% & 70\% \\ \hline
 Lower bound  & $ 0.834(0.015) $ & $ 0.845(0.005) $ & $ 0.864(0.004)$ & $0.722(0.011) $ & $0.746(0.011) $ & $ 0.772(0.009) $ & $ 0.613(0.004) $ & $0.634(0.004) $ & $ 0.648(0.012) $ \\ 
 CE  & $0.842(0.014) $ & $  0.858(0.003) $ &$  0.870(0.004) $ & $ 0.747(0.011) $ & $ 0.768(0.006) $ & $ 0.781(0.003) $ & $ 0.634(0.005) $ & $ 0.648(0.007) $ & $ 0.655(0.004)$  \\ 
 \textbf{BCE}  & $ \mathbf{0.851(0.013)} $ & $ 0.865(0.006) $ & $ 0.866(0.003) $ & $ 0.753(0.009) $ & $ 0.776(0.010) $ & $ 0.782(0.005) $ & $ 0.634(0.008) $ & $ 0.648(0.008) $ & $ 0.658(0.002)$ \\ 
 \textbf{GCE}  & $ 0.848(0.013) $ & $ \mathbf{0.865(0.005)} $ & $ 0.865(0.002) $ & $ \mathbf{0.756(0.001)} $ & $ \mathbf{0.779(0.007)} $ & $ \mathbf{0.786(0.004)} $ & $ 0.638(0.004) $ & $ \mathbf{0.653(0.001)} $ & $ \mathbf{0.662(0.003)} $ \\ 
 \textbf{SCE}  & $ 0.850s(0.016) $ & $ 0.858(0.006) $ & $ \mathbf{0.871(0.002)} $ & $ 0.758(0.015) $ &  $0.777(0.006) $ & $ 0.774(0.009) $ & $ \mathbf{0.641(0.010)} $ & $ 0.649(0.005) $ & $0.658(0.007)$ \\ \hline
 Upper bound &  \multicolumn{3}{c|}{$0.873$} & \multicolumn{3}{c|}{$0.788$} & \multicolumn{3}{c|}{ $0.668$ } \\ \hline
\end{tabular}
}
\label{tab:dsc}
\end{table}

%Subject one mainly shows the segmentation quality improvement on label 2 (yellow) and label 4 (blue), while for subject two it is label 1 (red).

\section{Conclusion}
We developed a semi-supervised learning strategy that uses ground-truth labels and generated pseudo-labels during training and applies robust loss functions to mitigate the negative effect on the model from noises existing in pseudo-labels. 
% A teacher-student framework is introduced. First, train a teacher model on labeled data and then generate pseudo labels for the unlabeled data. Next, train a student model on both ground-truth labels and pseudo labels with a robust loss function. We mainly investigated three robust losses in this work: BCE, GCE, and SCE. 
The proposed semi-supervised learning strategy is simple to deploy because of the plug-and-play robust loss module and has open possibilities for various applications as it is agnostic to specific model architecture.
The experimental results on classification and segmentation tasks show that the proposed semi-supervised learning strategy improved model performance, especially in scenarios where only a small amount of ground truth labels are available.

% Suggest a training strategy that uses both labeled  and additional unlabeled data for medical imaging problems. 
% First, use a trained network of labeled data only to generate a pseudo-one-hot label map of unlabeled  data. 
% These pseudo-labels are noisy, and the presence of noisy labels can disproportionately affecting the training process thus seriously impact the performance of classifiers. 
% Here we mitigate the effects of these noisy pseudo labels by using three robust loss functions. We show that this training strategy can improve DNN performance in classification, 
% especially segmentation, when the data-set is not extensively annotated.

%
\bibliographystyle{splncs04}
\bibliography{mybibliography}
\end{document}